\documentclass[letterpaper, 10 pt, conference]{ieeeconf} 
\IEEEoverridecommandlockouts                                                       

\usepackage{amsmath,amssymb,amsfonts}
\usepackage{algorithmic}
\usepackage{graphicx}
\usepackage{textcomp}
\usepackage{xcolor}
\usepackage{cuted}
\usepackage{capt-of}
\usepackage{gensymb}
\usepackage{booktabs}
\usepackage{multirow}
\usepackage{fancyhdr}
\def\BibTeX{{\rm B\kern-.05em{\sc i\kern-.025em b}\kern-.08em
    T\kern-.1667em\lower.7ex\hbox{E}\kern-.125emX}}

\usepackage[backend=biber,      %
    style=ieee,
    bibstyle=ieee,              %
    sortcites=true,   
    mincitenames=1,
    maxcitenames=3, %
    giveninits=true,            %
    backref=false
]{biblatex}
\renewcommand*{\bibfont}{\normalfont\small}
\usepackage{fancyhdr}
\newcommand{\mytitle}{\textbf{Submitted to appear in IEEE ICRA 2025.}}

\begin{document}

\title{ High Speed Robotic Table Tennis Swinging Using Lightweight Hardware with Model Predictive Control}

\author{David Nguyen$^{*}$, Kendrick D. Cancio$^{*}$, Sangbae Kim
    \thanks{$^{*}$Authors contributed equally. All authors are with the Biomimetic Robotics Laboratory at Massachusetts Institute of Technology (MIT).}
    \thanks{This work was supported by The AI Institute, Cambridge, MA.}
}

\maketitle

\thispagestyle{fancy}
\fancyhf{}		%
\fancyfoot[L]{\normalfont \sffamily  \scriptsize \mytitle}		%
\addtolength{\footskip}{-10pt}    %
\pagestyle{empty}

\begin{abstract}

    We present a robotic table tennis platform that achieves a variety of hit styles and ball-spins with high precision, power, and consistency. This is enabled by a custom lightweight, high-torque, low rotor inertia, five degree-of-freedom arm capable of high acceleration. To generate swing trajectories, we formulate an optimal control problem (OCP) that constrains the state of the paddle at the time of the strike. The terminal position is given by a predicted ball trajectory, and the terminal orientation and velocity of the paddle are chosen to match various possible styles of hits: loops (topspin), drives (flat), and chops (backspin). Finally, we construct a fixed-horizon model predictive controller (MPC) around this OCP to allow the hardware to quickly react to changes in the predicted ball trajectory. We validate on hardware that the system is capable of hitting balls with an average exit velocity of 11 m/s at an 88\% success rate across the three swing types.

\end{abstract}

\section{Introduction}

Robotic table tennis offers a compelling platform to take on many of the problems within the regime of dynamic manipulation where the object being handled is not stationary \cite{DynamicManipulation}. Unlike in other manipulation tasks, contact here must be intentional, impulsive, and performed with high speed and accuracy with the control of the manipulator and prediction of the ball's trajectory occurring simultaneously. Existing robotic table tennis systems that have tackled this generally suffer from two drawbacks. The first is that they may be specialized such as \cite{OMRON, WangDeltaLikeRobot} which require a large gantry to support a linkage mechanism that translates the paddle above the table. These systems cannot be easily modified to accomplish other dynamic manipulation tasks. Second, they may suffer from low acceleration due to their mass and rotor inertia. This is the case for \cite{KukaRobot, OnlineOptimalTraj, LiuStockArm} which utilize stock manipulators such as Kuka or WAM and makes it difficult for them to replicate faster strikes. In this paper, we present our lightweight, high-torque, low-inertia anthropomorphic arm as well as a model predictive control scheme and ball prediction model which together are capable of striking retro-reflective table tennis balls with various swing types. 

\begin{figure}
    \vspace{2mm}
    \centering
    \begin{minipage}{\linewidth}
        \centering
        \includegraphics[width=\linewidth]{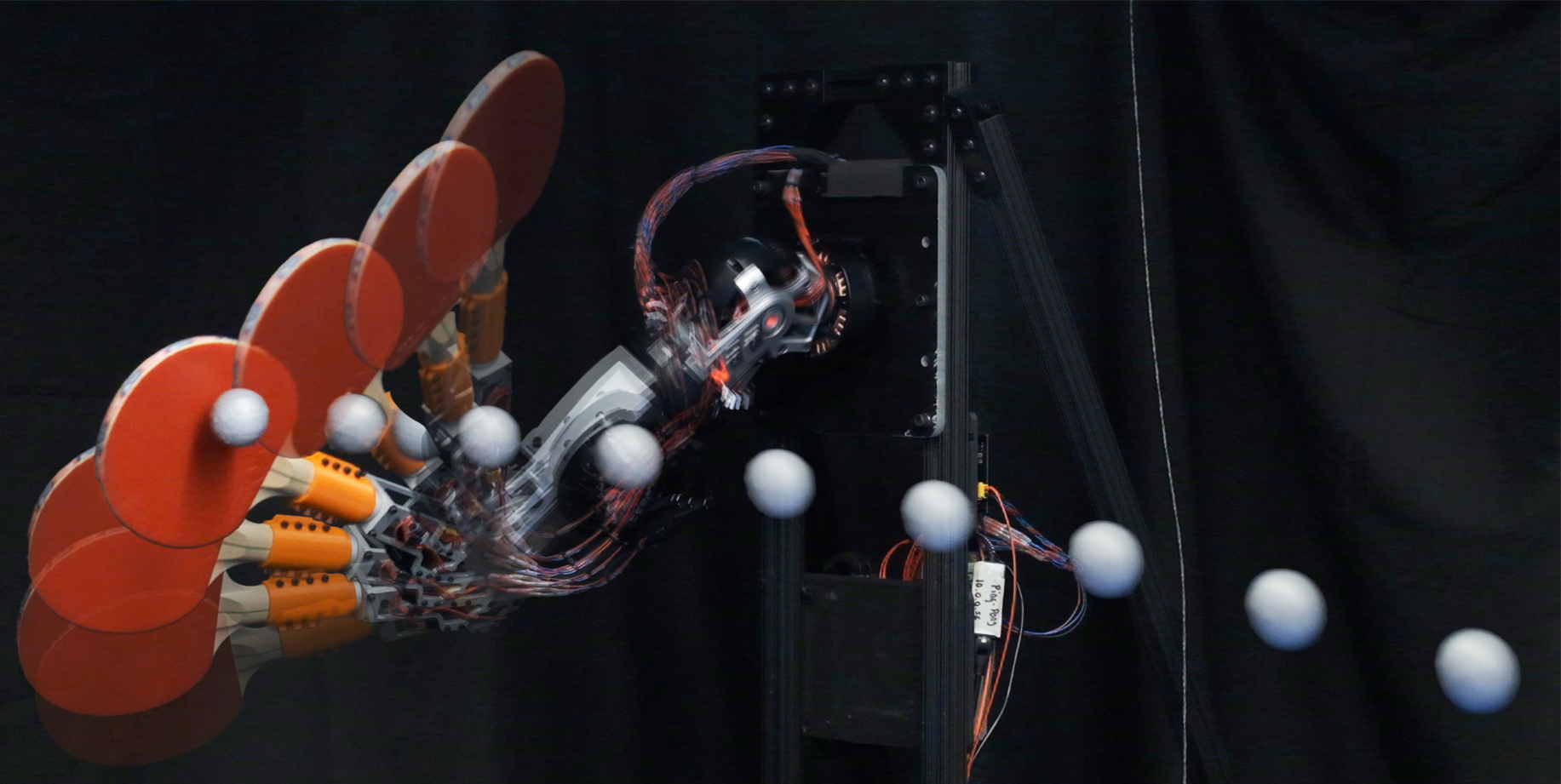}
    \end{minipage}
    
    \vspace{2mm} %
    
    \begin{minipage}{\linewidth}
        \centering
        \includegraphics[width=\linewidth]{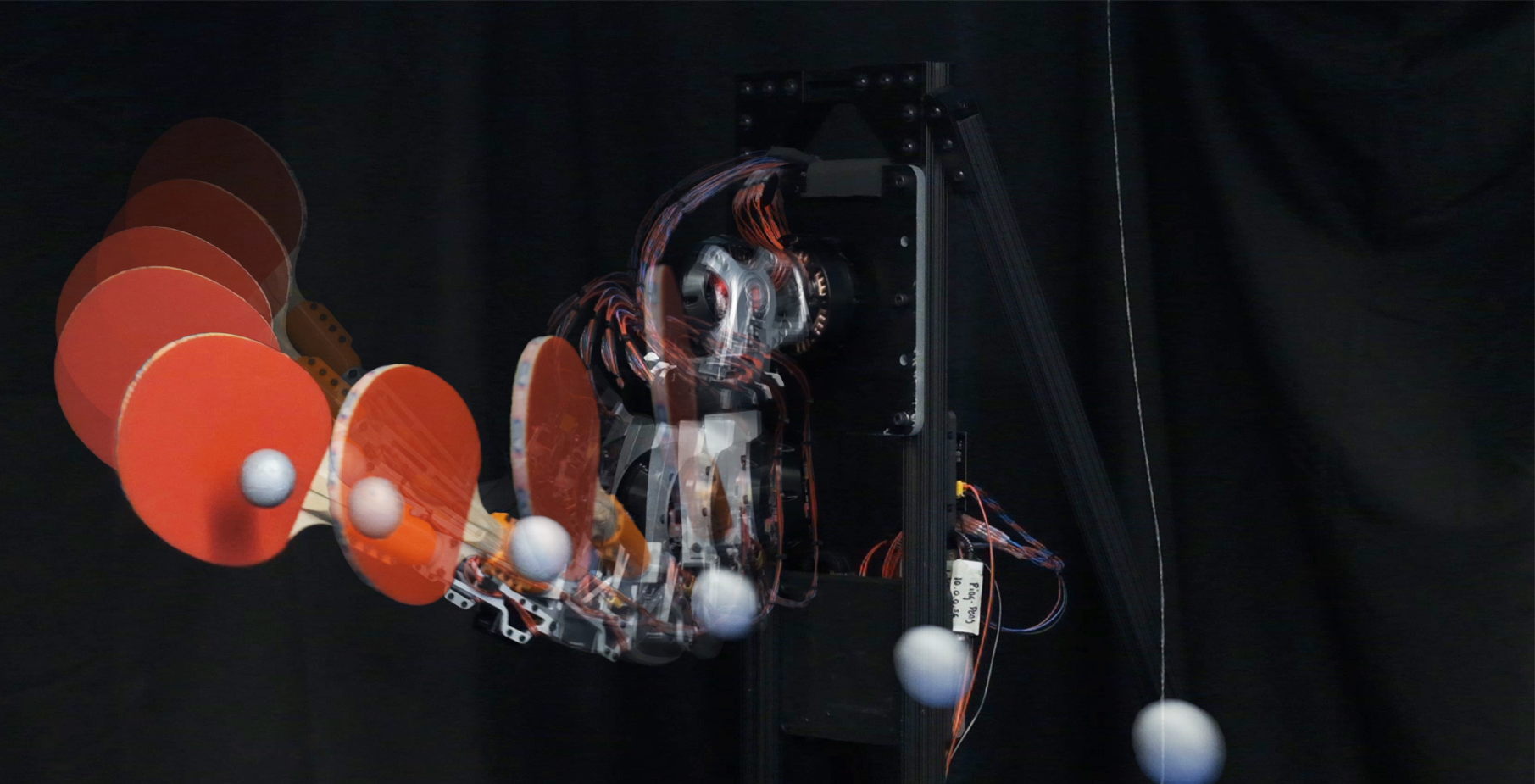}
    \end{minipage}
    
    \vspace{2mm} %
    
    \begin{minipage}{\linewidth}
        \centering
        \includegraphics[width=\linewidth]{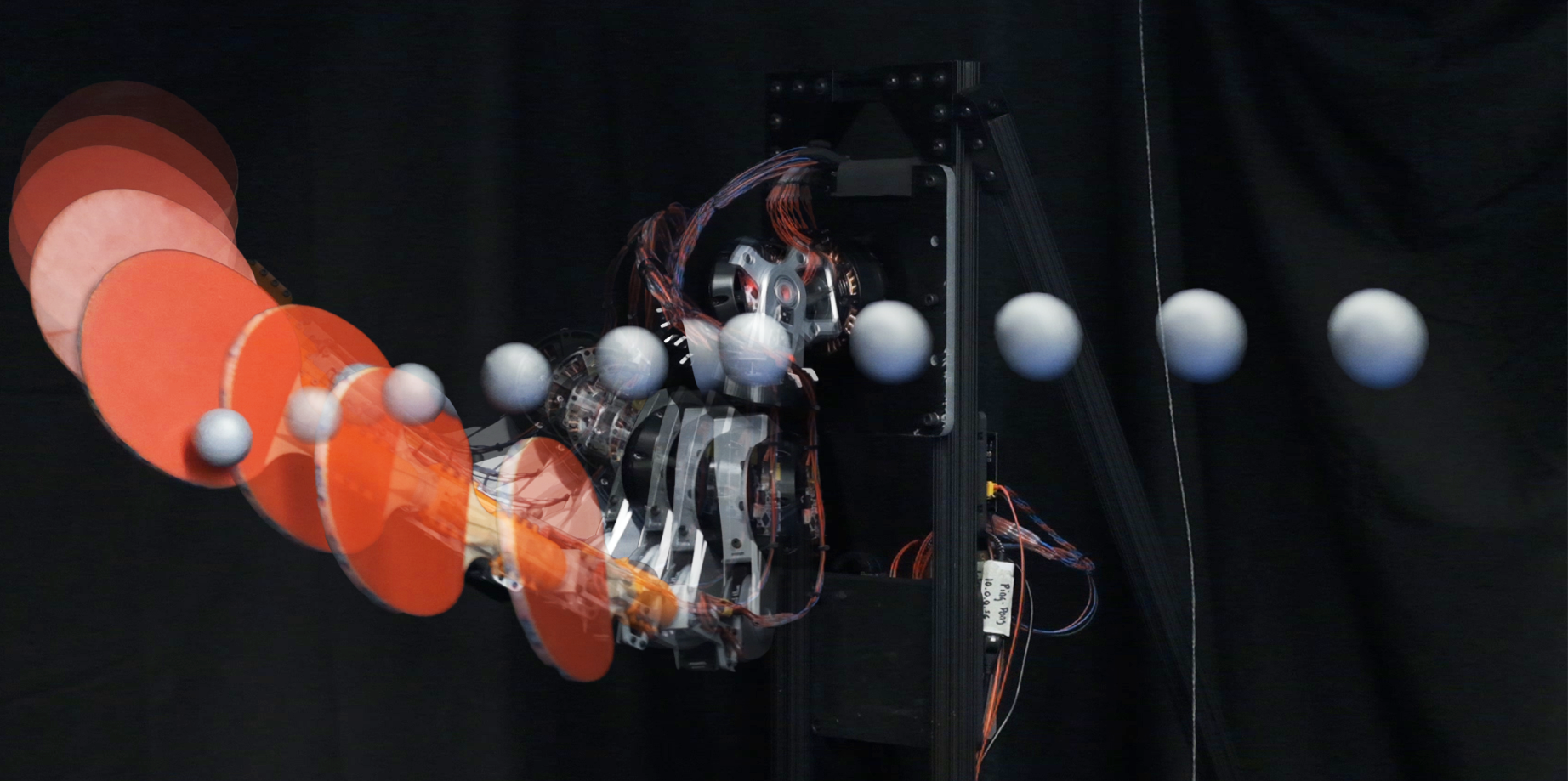}
    \end{minipage}
    
    \caption{\textbf{Hardware Strikes.} Loop (top), drive (middle), and chop (bottom) with exit ball trajectory shown.}
    \label{fig:swing_graphic}
\end{figure}

We focus on replicating forehand loops (top spin), drives (flat paddle), and chops (back spin) as seen in Fig.~\ref{fig:swing_graphic} with the ultimate goal of reaching human parity on these types of shots. Advanced human players are capable of returning balls using the forehand loop at about 21~m/s and drives at around 25~m/s \cite{HumanMetric1}. Robotic systems have struggled to reach this level of performance. To the best of our knowledge, \textcite{MuscleRobot} report the fastest strike speeds with average ball exit velocities of 12~m/s but with pneumatic actuation of a four degree-of-freedom (DoF) arm which limits the types of achievable swings. \textcite{BestGoogle}, who have a system capable of beating $55\%$ of intermediate human players, demonstrate a variety of swing types through their learned low level skill libraries but can only achieve exit velocities of at most 6.8~m/s. Using our custom hardware and high-bandwidth control system, we show progress towards matching the top end of human performance at striking table tennis balls while maintaining the ability to perform different types of shots.

\section{System}

\begin{figure}
    \centering
    \includegraphics[width=\linewidth]{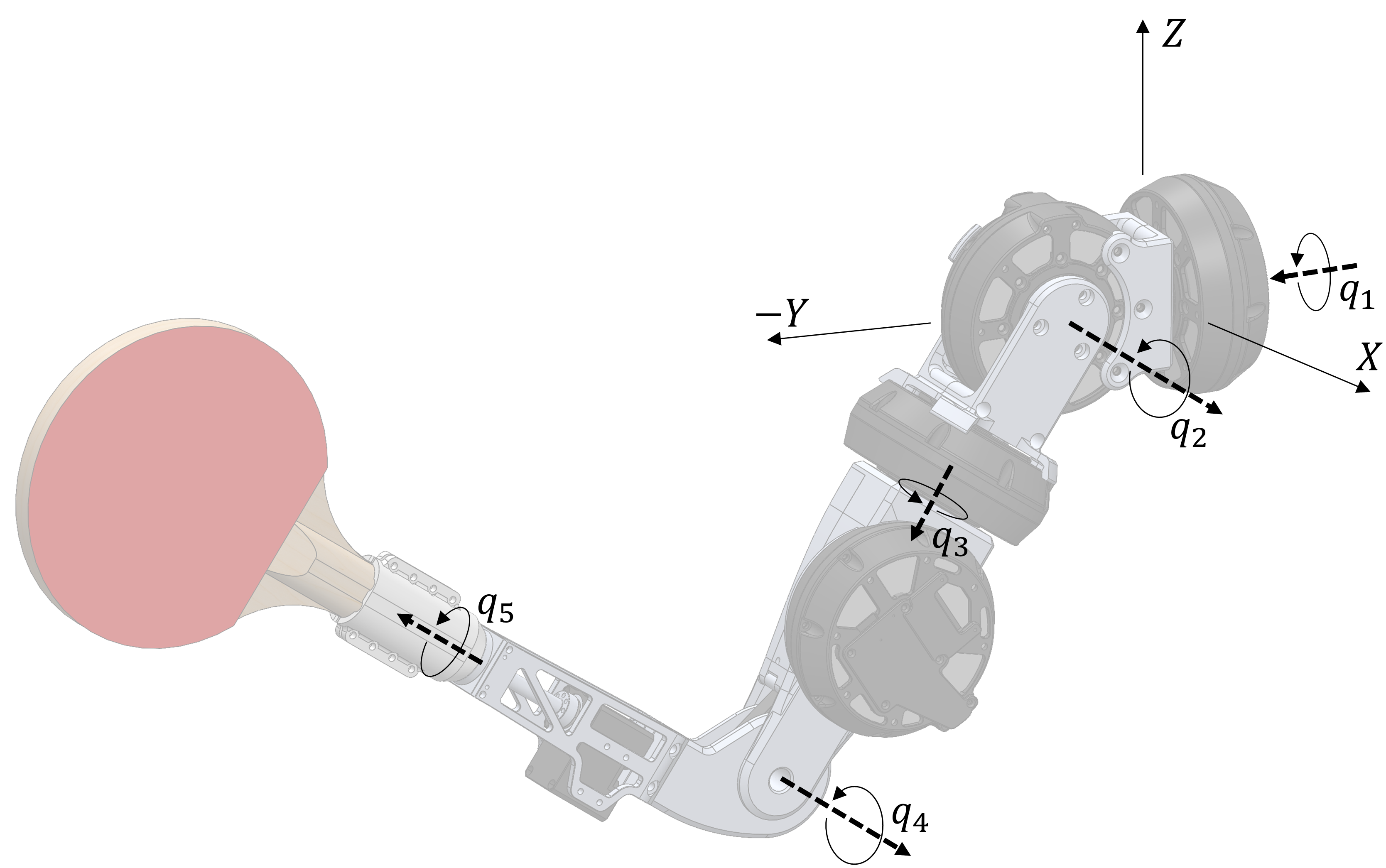}
    \caption{\textbf{Modified MIT Humanoid Arm.} The arm shoulder consists of $q_1$-$q_3$ with directly connected actuators while the $q_4$ actuator controls the elbow with a short belt transmission. $q_5$ is a smaller Dynamixel motor since it requires little torque to orient the paddle.}
    \label{fig:armdiagram}
\end{figure}

\begin{figure}
    \centering
    \includegraphics[width=\linewidth]{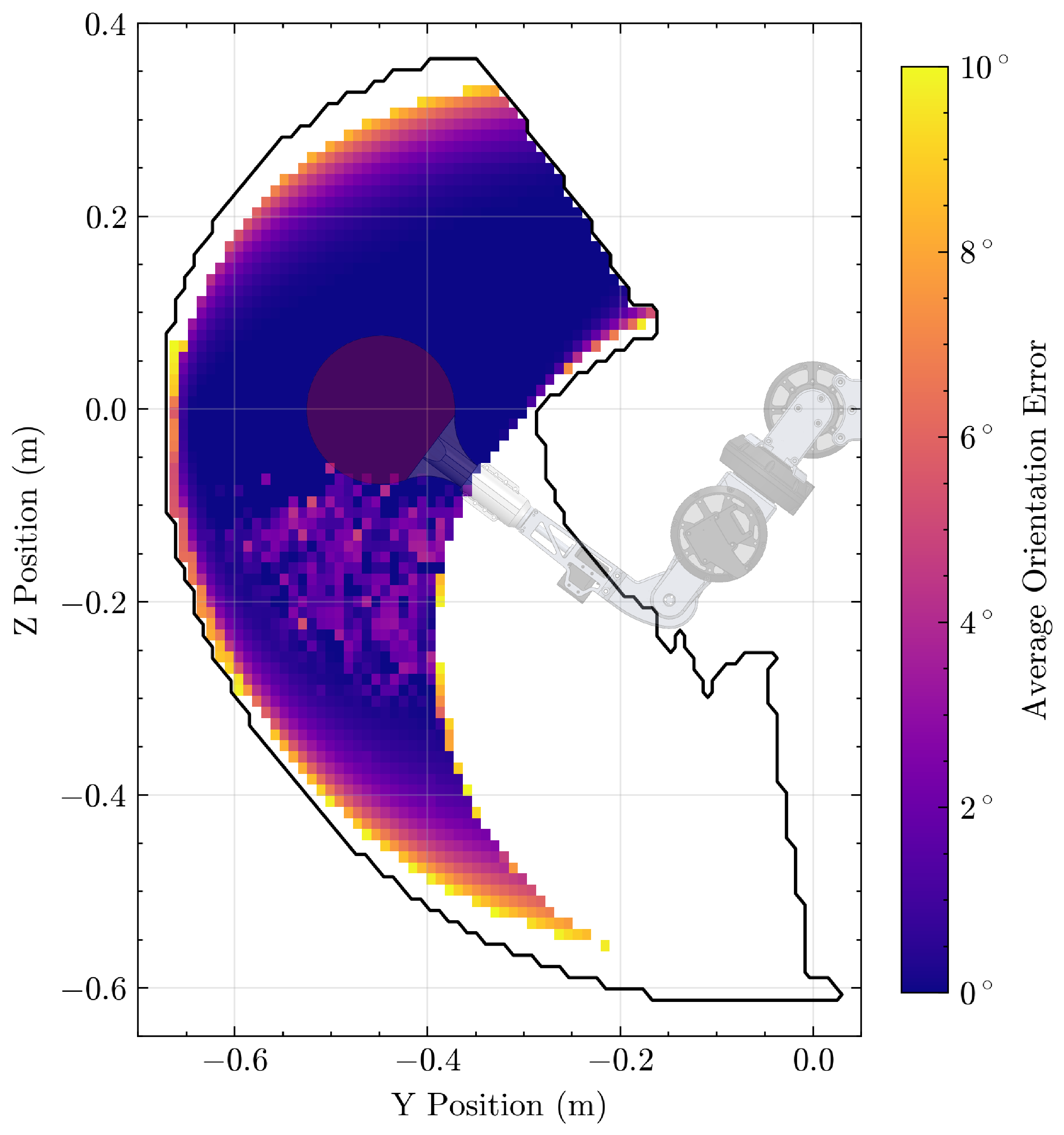}
    \caption{\textbf{Arm workspace analysis.} The black outline represents all achievable positions while the colored region is a heat map of the average orientation error bounded to 10\degree. The discolored region near $Y=-0.4$, $Z=-0.2$ is due to the wrist and first shoulder DoFs being close to singularity making some orientations difficult to reach.}
    \label{fig:workspace}
\end{figure}

\subsection{Hardware}
To achieve the speed and acceleration required for human-level table tennis swings, we use a modified 5 DoF version of the MIT Humanoid arm \cite{Humanoid}. As shown in Fig~\ref{fig:armdiagram}, the arm was designed with the actuator mass at the proximal end to reduce its inertia similar to the design principles used for high speed batting \cite{HighSpeedBatting}. The arm has a total mass of 3 kg with four U10 \cite{U10} actuators outputting 34 Nm of peak torque with 0.00612\ $\text{kg}\cdot\text{m}^2$ of effective rotor inertia. The 5th DoF is actuated by a Dynamixel weighing just 82 grams \cite{Dynamixel}, and is responsible for orienting the paddle and wrist. Because of the lightweight design, torque-dense actuation, and low reflected inertia, the arm is capable of acceleration in the range of 180-300 m/s$^2$ at the end effector. Together, these allow us to dictate the paddle's position (3 DoF) and surface normal (2 DoF) within our workspace. A 6th degree of freedom was omitted from the arm design since any paddle rotation about the paddle normal does not change the contact conditions.

To identify the area of feasible positions where the paddle could also reach ample orientations, we conducted a workspace analysis using an simple kinematic solver. This study was performed at the  $Y$-$Z$ plane intersecting the shoulder where the arm intends to strike the ball when swinging. We call this the strike-plane. The normal vector to the paddle surface was used to measure how well the paddle reached its desired orientation. At each tested position on the strike-plane, the desired normal of the paddle was swept between $\pm15\degree$ on the horizontal and $\pm45\degree$ on the vertical. The error for each tested paddle orientation was averaged for each strike-plane position, shown in Fig.~\ref{fig:workspace}. The large area of the arm's workspace where the orientation error is less than $10\degree$ indicates that it is capable of swinging in a variety of ways throughout its workspace.

We use six OptiTrack Flex 13 motion capture cameras to track standard table tennis balls wrapped in retro-reflective tape \cite{OptiTrack}. Although this alters the dynamics of the ball, this system was chosen as an off-the-shelf tracking solution with low latency, a 120 Hz frame rate, and sub-millimeter precision.

\subsection{Communication}

\begin{figure}
    \centering
    \includegraphics[width=\linewidth]{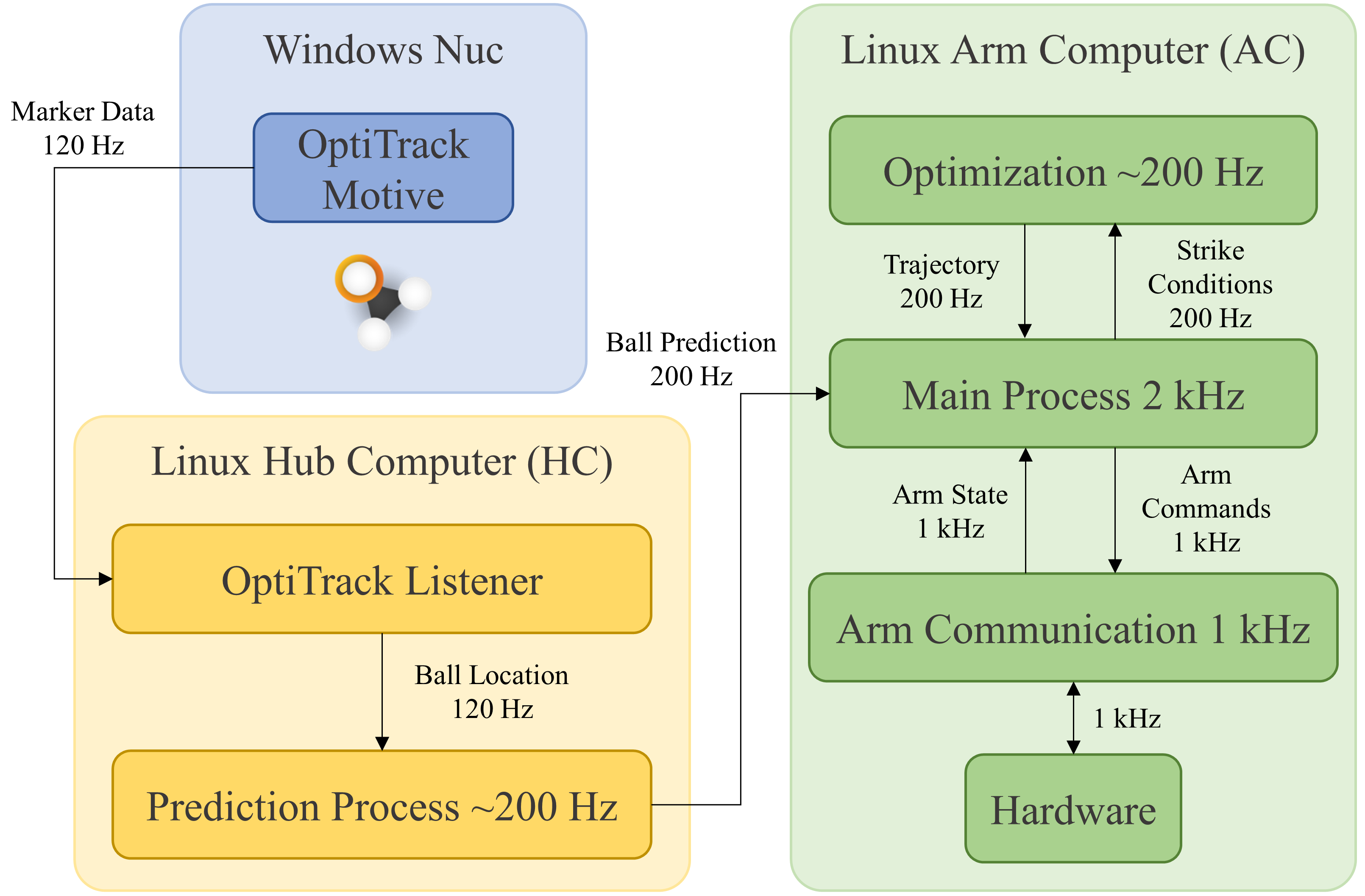}
    \caption{\textbf{System Communication Diagram.} In blue, the Windows machine runs the OptiTrack Motive software which is responsible for reading the cameras, capturing marker data, and publishing it over the NatNet network. A separate Linux Hub Computer (HC), in yellow, receives and filters the marker data then publishes it over LCM in the OptiTrack Listener process. Also on the HC, the Prediction Process computes the expected ball state as it reaches the arm which is then sent through LCM to a second Linux Arm Computer (AC). The AC, in green, receives the prediction in the Main Process and then interfaces with the Optimization Process to receive a new trajectory based on the strike conditions. Finally, the Main Process sends arm commands to the Arm Communication Process which handles low-level control and interfaces with the hardware via Ethernet and CAN on a middleware board.}
    \label{fig:system}
\end{figure}

The system consists of three computers running multiple processes that communicate via Lightweight Communications and Marshalling (LCM) \cite{LCM} and NatNet \cite{NatNet} over an Ethernet network as illustrated in Fig.~\ref{fig:system}. We aimed to minimize the effective reaction time, the time between a new ball observation and updated motor actuation. This includes communication latency, ball prediction time, and arm trajectory optimization time. These were measured to be 0.2-0.6 ms, 0.5-7.0 ms, and 4.5-6.5 ms respectively. In total, the effective reaction time between receiving new ball observations and executing new trajectories on hardware was between 7.5-16 ms.

\section{Ball Trajectory Prediction}
\subsection{Dynamics}

To predict the ball's trajectory when launched toward the arm, we integrate a simplified version of the dynamics \cite{SpinDetection} assuming that the unobserved incoming spin is zero and constant. The dynamics are shown in equations (\ref{eq:flightdynamics}) and (\ref{eq:bouncedynamics}) where $\mathbf{a}$ and $\mathbf{v}$ are the acceleration and velocity vectors of the ball respectively and $\mathbf{a_g}$ is the acceleration due to gravity. The term $\mathbf{v'}$ represents the post-bounce velocity and the coefficients $D$, $C_h$, and $C_v$ are the drag and restitution parameters fitted to observed ball trajectories.

\vspace{10mm}

\begin{equation}
\label{eq:flightdynamics}
\text{Flight Dynamics:} \quad 
\mathbf{a}_{Ball} = -D\|\mathbf{v}\|\mathbf{v} + \mathbf{a_g}
\end{equation}

\begin{equation}
\label{eq:bouncedynamics}
\text{Table Bounce:} \quad 
\begin{aligned}
v'_x &= C_h \cdot v_x \\
v'_y &= C_h \cdot v_y \\
v'_z &= -C_v \cdot v_z
\end{aligned}
\end{equation}

More explicitly, the $D$ term is a lumped drag parameter which encapsulates all the physical parameters of the ball and environment that contribute to the drag force on the ball. We estimate this term by performing a least squares linear fit for the relation $||\mathbf{a}_{Ball}-\mathbf{a_g}|| = D||\mathbf{v}||^2$ using the observed velocities and accelerations from collected ball trajectories.

The terms $C_v$ and $C_h$ relate the pre-bounce and post-bounce vertical and horizontal velocities of the ball and are also estimated using a least squares fit on observed pre and post-bounce velocities.

\subsection{State Estimation and Prediction}
The current state of the ball, consisting of its position and velocity, is used as the initial condition for the dynamics integration terminating at the strike-plane. We note this point on the strike-plane as $\mathbf{p_{des}}$ where the arm attempts to hit the ball at the accompanying strike-time $t_{strike}$. For estimating the state, the position of the ball is given directly by OptiTrack. The velocity is attained following the method in \cite{SpinDetection} where 3rd-order polynomials are fit to the observed positions and times of the ball. Evaluating the derivative of these polynomials at the current time gives a more accurate velocity estimate compared to finite difference methods. Because the arm controller directly responds to  $\mathbf{p_{des}}$, we reduce the variance between consecutive predictions by using a ten point moving average.

\begin{figure}
    \centering
    \includegraphics[width=\linewidth]{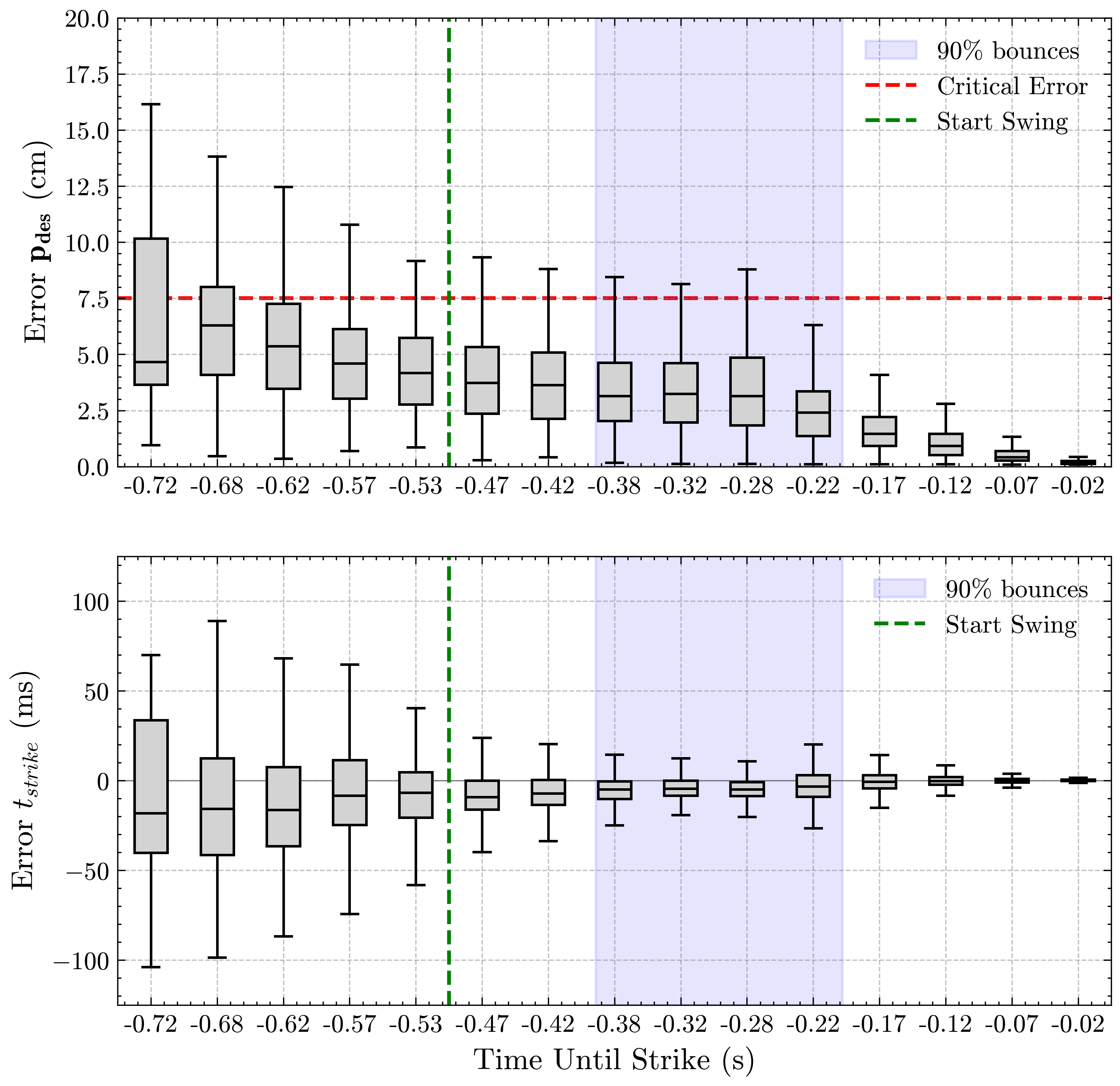}
    \caption{\textbf{Characterized Prediction Error.} The top and bottom plots show the distribution errors in $\mathbf{p_{des}}$ and $t_{strike}$ respectively as the ball approaches the strike-plane. The blue shaded region indicates the times where 90\% of the balls bounce, the green line marks when a 0.5 second swing begins, and the red line is the critical error of 7.5 cm of the paddle radius.}
    \label{fig:PredictionErrors}
\end{figure}

The prediction algorithm was evaluated by measuring $\mathbf{p_{des}}$ and $t_{strike}$ throughout the incoming ball trajectory and comparing them to their true values. Figure \ref{fig:PredictionErrors} shows that on average, the error for $\mathbf{p_{des}}$ is well within half the width of the racket by the time the swing is initiated.
These errors sharply diminish after the ball bounces on the table which occurs on average 0.25 seconds before the strike. Our model also consistently predicts $t_{strike}$ within 0.25 ms after the swing is initiated.

\section{Arm Controller}

To generate swinging trajectories for the arm, we formulated the task as an optimal control problem (OCP) which enables us to handle the kinematic limitations of the arm and the strike conditions of the paddle as constraints. This OCP is implemented in a Model Predictive Controller (MPC) that re-plans the trajectories given the most updated state information to handle changes in the strike conditions during a swing.

Equations \eqref{eq:cost}-\eqref{eq:paddle_orientation_constraint} outline the OCP used in the MPC where parameter inputs $\mathbf{p_{des}}$, $\mathbf{v_{des}}$ and $\mathbf{o_{des}} \in \mathbb{R}^3$ define the terminal position, velocity, and orientation of the paddle when hitting the ball. The cost function minimizes the total acceleration and velocities of each node weighted by $w_a$ and $w_v$ respectively \eqref{eq:cost}. The initial conditions of the trajectory are constrained by parameters $\mathbf{q_0}$ and $\mathbf{\dot{q}_0}$ $\in \mathbb{R}^5$. 

\begin{figure*}[h]
    \centering
    \includegraphics[width=\linewidth]{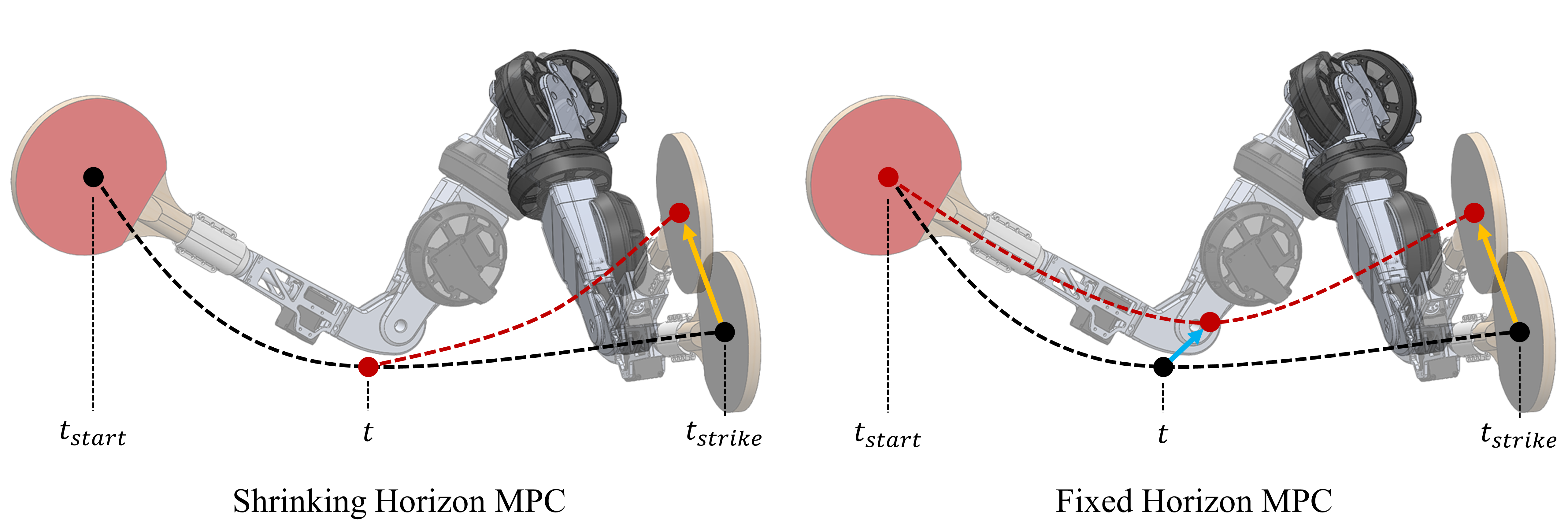}
    \caption{A diagram of the traditional Shrinking Horizon MPC formulation (left) compared to our Fixed Horizon formulation (right). The orange arrows represents a change in terminal strike conditions with the black lines being the prior solution and the red lines being the new solution. On the right the blue arrow represents the shift between trajectories in the Fixed Horizon MPC. The trajectory start time is represented by $t_{start}$, the current time by $t$, and the time at which the strike occurs by $t_{strike}$.}
    \label{fig:mpc_diagram}
\end{figure*}

\begin{subequations}
\label{optimization}
\begin{align}
\min_{\mathbf{q[.], \dot{q}[.], \ddot{q}[.]}} \quad & \sum_{n=0}^{N} w_a\mathbf{\ddot{q}}[n]^\intercal\mathbf{\ddot{q}}[n] + w_v\mathbf{\dot{q}}[n]^\intercal\mathbf{\dot{q}}[n] \label{eq:cost}\\
\textrm{s.t.} \quad & \mathbf{q}[0] = \mathbf{q_0} \label{eq:init_pose} \\
  & \mathbf{\dot{q}}[0] = \mathbf{\dot{q}_0} \label{eq:init_velocity} \\
  & \mathbf{q_{min}} <= \mathbf{q}[n] <= \mathbf{q_{max}} \quad \forall{n} \label{eq:joint_limits} \\
  & \mathbf{q}[n+1] = \mathbf{q}[n] + \mathbf{\dot{q}}[n]\Delta t \quad \forall{n} \label{eq:vel_constraint} \\
  & \mathbf{\dot{q}}[n+1] = \mathbf{\dot{q}}[n] + \mathbf{\ddot{q}}[n]\Delta t \quad \forall{n} \label{eq:acc_constraint} \\
  & ||\operatorname{FK}(\mathbf{q_f}) - \mathbf{p_{des}}||_2^2 <= \epsilon_p \label{eq:paddle_position_constraint} \\
  & ||J(\mathbf{q_f})\mathbf{\dot{q}_f} - \mathbf{v_{des}}||_2^2 <= \epsilon_v \label{eq:paddle_velocity_constgraint} \\
  & ||\operatorname{FK_o}(\mathbf{q_f}) - \operatorname{FK}(\mathbf{q_f}) - \mathbf{o_{des}}||_2^2 <= \epsilon_o \label{eq:paddle_orientation_constraint}
\end{align}
\end{subequations}

The joint position, velocity, and acceleration trajectory matrices are $\mathbf{q}[.]$, $\mathbf{\dot{q}[.]}$, $\mathbf{\ddot{q}}[.] \in \mathbb{R}^{5 \times N}$ respectively. Equations \eqref{eq:joint_limits}-\eqref{eq:acc_constraint} include the joint limit ($\mathbf{q_{max}}$ and $\mathbf{q_{min}}$) and dynamics constraints with $\Delta t$ as the time step. The position and velocity vectors of the end effector are obtained using the forward kinematics function $\operatorname{FK}$ and Jacobian of the paddle center $J$, at $\mathbf{q_f}$ the final position node $\mathbf{q}[N]$. These values are constrained to be within tolerances $\epsilon_p$ and $\epsilon_v$.

We define orientation as the unit normal vector from the face of the paddle: $\operatorname{FK_o}(\mathbf{q_f}) - \operatorname{FK}(\mathbf{q_f})$ where $\operatorname{FK_o}(\mathbf{q_f})$ is the forward kinematics function for the normal vector point above the paddle face. To simplify the OCP, the distance metric between orientation vectors is defined by their Euclidean distance rather than intersecting angle. Therefore equation \eqref{eq:paddle_orientation_constraint} constrains the paddle orientation to be within $\epsilon_o$ of the desired unit normal vector $\mathbf{o_{des}}$. This optimization problem was constructed using CasADi \cite{CASADI} and the solver used was IPOPT \cite{IPOPT}.

\subsection{Model Predictive Control Implementation}

There are two ways to implement this OCP in a model predictive controller, the first uses the current state of the arm as inputs for $\mathbf{q_0}$ and $\mathbf{\dot{q}_0}$ to seed the problem. This requires the $\Delta t$ term in the OCP to shrink as the arm approaches $\mathbf{p_{des}}$. We call this implementation Shrinking Horizon MPC (SHMPC) which is a special case of the Variable Horizon MPC \cite{VariableMPC}. The second approach uses the ready state of the arm as the initial conditions rather than current state while keeping the $\Delta t$ constant. This is equivalent to optimizing for the full swing from the ready position only adjusting for the updated $\mathbf{p_{des}}$. We call this method Fixed Horizon MPC (FHMPC). A drawback to FHMPC is that subsequent new solutions are not guaranteed to be close to the current state which can result in aggressive set point changes to track the new trajectory. Illustrations of both SHMPC and FHMPC are shown in Fig.~\ref{fig:mpc_diagram}. Both implementations use their prior solutions to warm start the next OCP to accelerate convergence and reduce solve time.

We compare these two MPC implementations by simulating them using fixed strike conditions as well as ones captured from 82 trials of real-world prediction data. We also compared MPC performance without warm starting on the same prediction data. Table \ref{tbl:mpc_eval} shows the results of the evaluation of the three tests. The convergence ratio was calculated by comparing the number of converged solutions to the total number of attempted solves. The results demonstrate that FHMPC solves twice as fast and converges more often than SHMPC. For the tests where the implementations were not warm started, solve time was slightly faster for SHMPC which indicates that FHMPC warm starts more effectively. We can attribute this to the changing timescale of SHMPC where each solution is not necessarily a good initial guess for the next shorter problem.  

\begin{table}
\caption{MPC Implementation Comparison}
\begin{center}
\begin{tabular}{c c c c c c}
& & \multicolumn{2}{c}{\textbf{FHMPC}}&\multicolumn{2}{c}{\textbf{SHMPC}} \\
\cmidrule(lr){3-4} \cmidrule(lr){5-6}
\textbf{Ball} & \textbf{Warm} & \textbf{\textit{Converge}}& \textbf{\textit{Solve}}& 
\textbf{\textit{Converge}}& \textbf{\textit{Solve}} \\
\textbf{Prediction}& \textbf{Start} & \textbf{\textit{Ratio}}& \textbf{\textit{Time$^{\mathrm{1}}$}}& \textbf{\textit{Ratio}}& \textbf{\textit{Time$^{\mathrm{1}}$}} \\
\toprule
Fixed&Yes& \textbf{99.6\%} & \textbf{2.8 ms} & 89.6\% & 6.2 ms \cr
Data&Yes& \textbf{99.5\%} & \textbf{3.2 ms} & 91.2\% & 6.7 ms \cr
Data&No& 65.1\% & 25.2 ms & 66.3\% & 23.7 ms \cr
\bottomrule
\multicolumn{6}{l}{$^{\mathrm{1}}$Solve times from a computer with an Intel i9-14900K CPU.}
\end{tabular}
\label{tbl:mpc_eval}
\end{center}
\end{table}

Given the data, we chose FHMPC because it provides more solutions during a swing with its faster solve time and high convergence rate. This makes the arm more reactive to new ball predictions. To address the drawbacks of FHMPC, our prediction algorithm carries a moving average of $\mathbf{p_{des}}$ which in turn reduces the variance between subsequent solutions from FHMPC. To handle large set point changes, we also smoothly transition from the prior trajectory to the new solution using an S-curve over 20 ms. This short transition is enabled by our hardware which can handle high accelerations.

Finally, to ensure the arm meets the ball at the appropriate strike time $t_s$, we select the index $i^*$ along the optimal trajectory  which corresponds to the remaining time left in the swing. Equations \eqref{eq:index} and \eqref{eq:index_clip} outline this logic which allows the arm to jump forward or slow down along the trajectory for correct timing. Here $t$, $T_{swing}$ are the current time, and the swing duration respectively while $N_i$ is the number of nodes after interpolation and $S_{max}$ is the maximum step the MPC can jump forward in the trajectory.

\begin{subequations}
    \begin{align}
        i = (1-\frac{t_{strike} - t}{T_{swing}})*N_i \label{eq:index} \\
        i^* = \text{min}(\text{max}(i,0),S_{max}) \label{eq:index_clip}
    \end{align}
\end{subequations}

\subsection{Low-Level Control}

The output of the MPC formulation is a state and acceleration trajectory that is then interpolated to get nodes at 2 ms intervals. Since the arm system can be treated as fully actuated, the dynamics shown by \eqref{eq:dynamics} can be inverted to get a feed forward torque, $u$ in \eqref{eq:feedback_lin} based on the state and desired acceleration, $\ddot{q}_{des}$ from the optimized trajectory.

\begin{subequations}
\begin{align}
    M(\mathbf{q})\mathbf{\ddot{q}} + C(\mathbf{q},\mathbf{\dot{q}}) = \tau_g(\mathbf{q}) + \mathbf{u} \label{eq:dynamics} \\
    \mathbf{u} = M(\mathbf{q})\mathbf{\ddot{q}_{des}} + C(\mathbf{q},\mathbf{\dot{q}}) - \tau_g(\mathbf{q}) \label{eq:feedback_lin}
\end{align}
\end{subequations}

In \eqref{eq:dynamics} and \eqref{eq:feedback_lin} $M(\mathbf{q})$, $C(\mathbf{q}, \mathbf{\dot{q}}$) and $\tau_g(\mathbf{q})$ represent the mass matrix, Coriolis vector, and gravity vector respectively. Together the resulting $\mathbf{u}$, along with a simple joint PD controller, ensures the arm joints track the trajectory set points accurately.

\section{System Evaluation}

\begin{figure}
    \centering
    \includegraphics[width=0.7\linewidth]{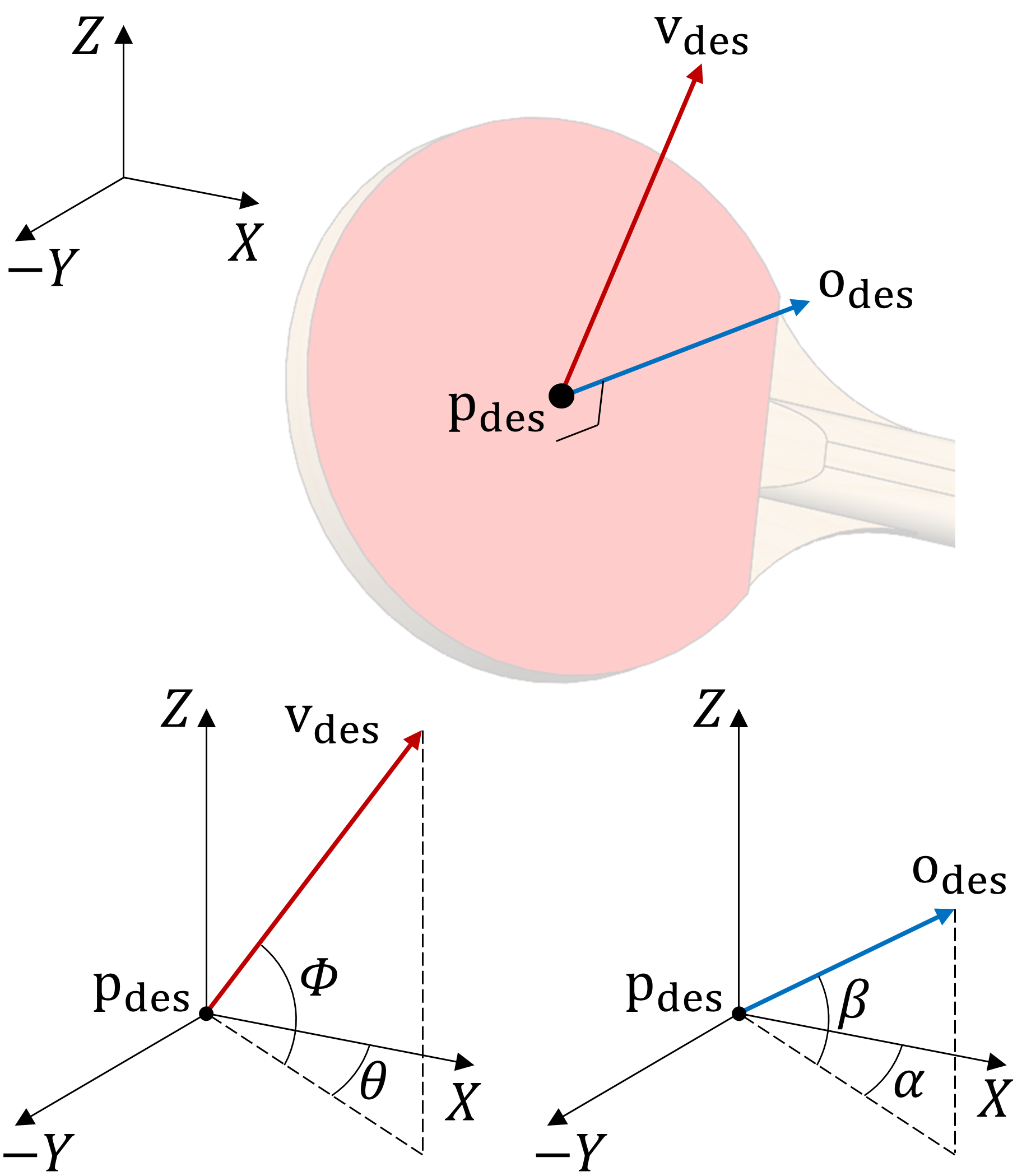}
    \caption{\textbf{Velocity and Orientation Angle Definition.} The orientation vector $\mathbf{o_{des}}$ is normal to the paddle surface and not necessarily parallel to $\mathbf{v_{des}}$. Velocity is parameterized by $\phi$ and $\theta$ and orientation defined by $\beta$ and $\alpha$ on the bottom two diagrams. ${XYZ}$ is the world frame.}
    \label{fig:velocity_orientation}
\end{figure}

The system we have presented is able to strike balls launched into the arm's workspace with the prescribed strike conditions. To holistically evaluate this capability, we performed trials with a variety of swing types to test the prediction, swing generation, and swing execution together on hardware. The tests included throwing 150 balls at the arm where the swing duration was set to 0.5 seconds and the terminal paddle velocity was set to 6 m/s. Table \ref{tbl:swings} shows the velocity and orientation parameters used for each swing type: loop, chop, and drive. The direction of the paddle velocity and orientation is defined by $\phi$, $\theta$, $\beta$, and $\alpha$ in Fig.~\ref{fig:velocity_orientation}.

\begin{table}
\caption{Swing Type Strike Conditions}
\begin{center}
\begin{tabular}{c c c c c}
\textbf{Swing Type} & $\mathbf{\theta}$ & $\mathbf{\phi}$ & $\mathbf{\alpha}$ & $\mathbf{\beta}$\\
\toprule
Loop & 0\degree & 45\degree & 0\degree & -7\degree\cr
Chop & 0\degree & -18\degree & 0\degree & 12\degree \cr
Drive & 0\degree & 0\degree & 0\degree & 0\degree \cr
\bottomrule
\end{tabular}
\label{tbl:swings}
\end{center}
\end{table}

\begin{figure}
    \centering
    \includegraphics[width=\linewidth]{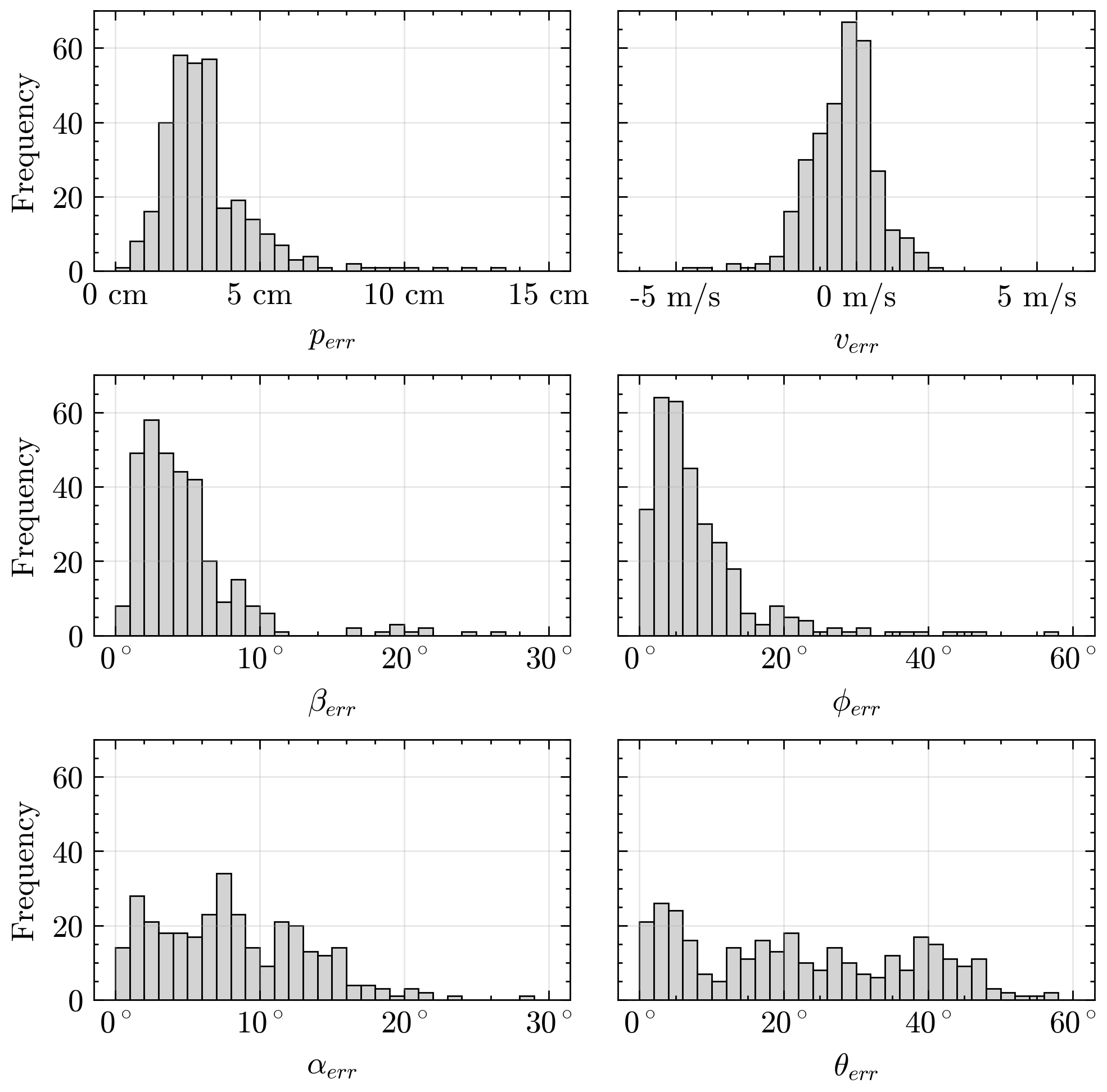}
    \caption{\textbf{Strike Condition Tracking.} The top left plot shows a histogram of $p_{err}$ which is the Euclidean distance between the measured paddle location and $\mathbf{p_{des}}$. The $v_{err}$ histogram on the top right indicates the difference between the measured velocity and $\mathbf{v_{des}}$ in magnitude. The bottom four histograms all correspond to $\beta$, $\alpha$, $\phi$, and $\theta$ in Fig.~\ref{fig:velocity_orientation} and their error is the absolute value of the angle difference between measured and desired.}
    \label{fig:tracking_error}
\end{figure}

For each test, the true strike conditions were measured ($\mathbf{p}$, $\mathbf{v}$, $\mathbf{o}$) along with the pre-strike and post-strike ball states. Out of the 150 balls thrown for each swing type, the hit rate was 88.4\% for loop strikes, 89.2\% for chops, and 87.5\% for drives. Histograms of the position, velocity magnitude, orientation, and velocity direction errors are shown in Fig.~\ref{fig:tracking_error} to illustrate how well the system reaches the desired strike conditions on hardware.

The position error is well within the 7.5 cm critical distance for a majority of strikes while the velocity magnitude is also within 2 m/s of the desired strike speed. For the paddle orientation, $\phi$ is within $10\degree$ of the desired direction whereas there is a higher variance when trying to achieve the desired $\theta$ with errors of up to 20\degree. Similarly, $v$ was more accurate on the vertical angle $\beta$ than on the horizontal angle $\alpha$. 

Overall, the variance in the velocity error distributions was higher than their orientation error counterparts. The large difference in the errors between $\phi$ and $\beta$ compared to $\theta$ and $\alpha$ can be explained by the lack of control authority in that dimension. Since the wrist's degree of freedom aligns closest with the y-axis when the ball is hit, small adjustments can be easily made to $\phi$ and $\beta$ while larger motions are required for changes in $\theta$ and $\alpha$.

\begin{figure}
    \centering
    \includegraphics[width=\linewidth]{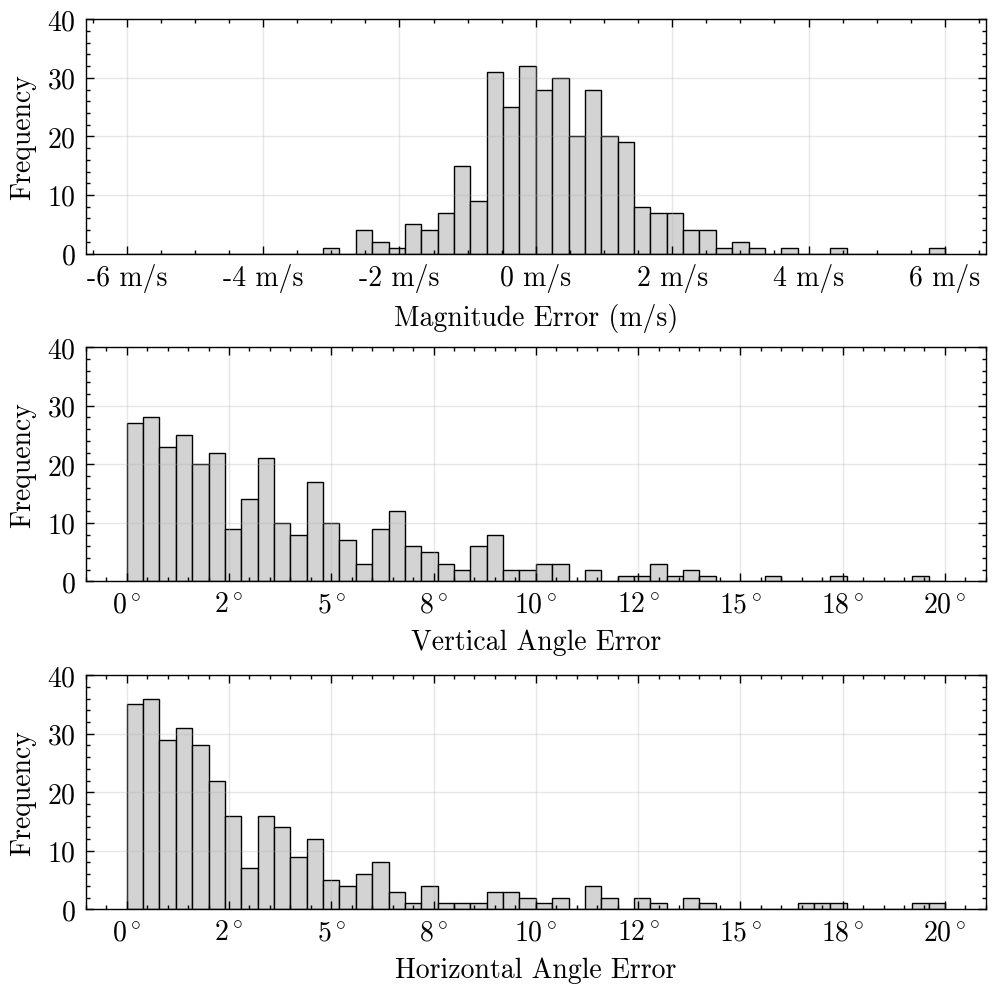}
    \caption{\textbf{Post-Strike Ball Velocity Error.} Histograms of the magnitude and angle errors between the predicted and measured exit ball velocity.}
    \label{fig:strike_error}
\end{figure}

To see how these errors impact the exiting state of the ball, the incoming ball velocity $\mathbf{v_{ball}^-}$ was used in a collision model from \cite{OnlineOptimalTraj} along with the target paddle state ($\mathbf{p_{des}}$, $\mathbf{v_{des}}$, and $\mathbf{o_{des}}$) to get a predicted exit ball velocity post-collision $\mathbf{v_{ball}^+}$. The contact model was modified to exclude the incoming spin of the ball and the contact parameters were tuned to match the hardware data as best as possible. The magnitude along with vertical and horizontal angles were compared between the predicted and the measured $\mathbf{v_{ball}^+}$. Distributions of these three errors are shown in Fig.~\ref{fig:strike_error}.

Similar to the paddle velocity tracking, a majority of the measured $\mathbf{v_{ball}^+}$ of the ball were within 2 m/s of the predicted value. For the vertical and horizontal directions, a majority of the exiting velocities were within $10\degree$ of the target which is far better than the paddle tracking error. The exiting velocities of the ball were between 7 and 11 m/s depending on the strike type. When testing the upper limits of our actuation, average ball velocities of up to 14 m/s were recorded with strikes where $|\mathbf{v_{des}}|$ was set to 10 m/s.

\section{Conclusion}

Our results show our robotic arm is capable of strike speeds that can match or exceed other robotic platforms while still retaining high hit rates. We do this while replicating three types of swings that vary the contact angle and velocity of the paddle when striking the ball and report return speeds of up to 14 m/s on average.

Although the peak capability of our system is promising, there are three drawbacks to our implementation. First, when generating trajectories, the current arm state is not guaranteed to be close to the new solution which can result in a large jump to track the new trajectory. Although this new solution is optimal for reaching the strike conditions from the ready state, it may be suboptimal to track from the current arm state. This is a fundamental drawback of the FHMPC over the SHMPC implementation. From the tracking results in Fig.~\ref{fig:tracking_error}, the arm also requires additional control authority at the wrist to control angles $\theta$ and $\alpha$ to strike more accurately. With an added wrist DoF, we would also no longer be constrained to the strike-plane allowing the system to choose $\mathbf{p_{des}}$ in three dimensions like \cite{OnlineOptimalTraj}. Finally, the prediction system has high latency variability from the changing number of instructions as the ball approaches the strike-plane. This alongside space constraints restricted the incoming ball speeds and was not a variable that we tested in our validation. Addressing all of these issues would enable the arm to better reach its terminal strike conditions for varying incomming ball speeds.

Throughout this work, our focus was on achieving dynamic, human-like striking. As such, our analysis is limited from the start of the swing to the moment the ball is struck. Aiming is not done explicitly but is shown to be possible by manually iterating over various strike conditions. Future work will focus on automating shot aiming through the use of a racket-ball collision model, expanding the workspace of the robot to reach the full game play area, and implementing RGB tracking to allow the use of regulation table tennis balls. Future work will also need to include spin estimation for both prediction of the ball's flight and collision trajectories. Together, these additional changes will bring the system closer to playing a game of competitive table tennis.

\section{Acknowledgements}

The authors would like to thank Se Hwan Jeon, Elijah Stanger-Jones, and Ronak Roy.

\printbibliography

\end{document}